\documentclass[journal]{IEEEtran}
\ifCLASSINFOpdf
\else
\fi
\usepackage{graphicx}
\usepackage{url,hyperref,lineno,microtype,subcaption,tabularx,xcolor}
\usepackage{mwe}
\usepackage{float}
\usepackage[onehalfspacing]{setspace}
\usepackage{amsmath}
\usepackage{amssymb}
\usepackage{bbm}
\usepackage{comment}
\usepackage[linesnumbered,ruled,vlined]{algorithm2e}
\DeclareMathOperator{\E}{\mathbb{E}}
\newcolumntype{M}[1]{>{\centering\arraybackslash}m{#1}}
\newcommand\algocomment[1]{%
  {\textcolor{blue}{\fontfamily{qcr}\selectfont\scriptsize{\textit{(#1)}}}}%
}
\newcommand\algocommentblock[1]{%
  {\textcolor{blue}{\fontfamily{qcr}\selectfont\scriptsize{\textbf{(#1)}}}}%
}

\begin{document}
%
\title{Autonomous learning of multiple, \\ context-dependent tasks}
%
%
%

\author{Vieri Giuliano Santucci, Davide Montella, Bruno Castro da Silva and Gianluca Baldassarre
\thanks{V.G. Santucci, D. Montella and G. Baldassarre are with Istituto di Scienze e Tecnologie della Cognizione (ISTC), Consiglio Nazionale delle Ricerce (CNR), Rome, Italy. B.C. da Silva is with Institute of Informatics, Federal University of Rio Grande do Sul (UFRGS), Porto Alegre, Brazil}%
\thanks{Corresponding author: V.G. Santucci, vieri.santucci@istc.cnr.it}
}

\maketitle

\begin{abstract}
When facing the problem of autonomously learning multiple tasks with reinforcement learning systems, researchers typically focus on solutions where just one parameterised policy per task is sufficient to solve them. 
However, in complex environments presenting different contexts, the same task might need a set of different skills to be solved.
These situations pose two challenges:
(a) to recognise the different contexts that need different policies;
(b) quickly learn the policies to accomplish the same tasks in the new discovered contexts. 
These two challenges are even harder if faced within an open-ended learning framework where an agent has to autonomously discover the goals that it might accomplish in a given environment, and also to learn the motor skills to accomplish them.  
We propose a novel open-ended learning robot architecture, C-GRAIL, that solves the two challenges in an integrated fashion.
In particular, the architecture is able to detect new relevant contests, and ignore irrelevant ones, on the basis of the decrease of the expected performance for a given goal.
Moreover, the architecture can quickly learn the policies for the new contexts by exploiting transfer learning importing knowledge from already acquired policies.
The architecture is tested in a simulated robotic environment involving a robot that autonomously learns to reach relevant target objects in the presence of multiple obstacles generating several different obstacles.
The proposed architecture outperforms other models not using the proposed autonomous context-discovery and transfer-learning mechanisms.
\end{abstract}

\begin{IEEEkeywords}
Autonomous robotics, Context-dependent tasks, Intrinsic Motivations, Reinforcement learning, Developmental Robotics.
\end{IEEEkeywords}

%
\IEEEpeerreviewmaketitle

\section{Introduction}
\label{sec:Intro}
%
%
%
%
\IEEEPARstart{I}{n} recent years, the development of autonomous agents has gained increasing interest in the fields of artificial intelligence, robotics, and machine learning. Although autonomy and versatility are pursued via different kinds of approaches such as information theory \cite{Klyubin2008,Martius2013}, evolutionary computation \cite{Lehman2011}, deep learning \cite{Gu2017deep}, or epigenetic models \cite{Lones2017}, the field of developmental robotics \cite{Cangelosi2015Book}, and in particular the research on intrinsically motivated open-ended learning \cite{Baldassarre2013Book,Santucci2020Editorial}, is producing a growing number of promising and cumulative results.

The concept of Intrinsic Motivations (IMs) is borrowed from biological \cite{White1959} and psychological literature \cite{Ryan2000} describing how novel or unexpected ``neutral'' stimuli, as well as the perception of control over the environment, can generate learning processes even in the absence of assigned rewards or tasks. In the computational literature, IMs have been implemented in artificial agents to foster their autonomy in gathering knowledge \cite{Schmidhuber2010,Hester2017}, learning repertoire of skills \cite{Chentanez2005,Schembri2007a,Metzen2013,Oudeyer2013}, exploiting affordances from the environment \cite{Hart2013,Baldassarre2019Afford,Manoury2019}, selecting their own tasks \cite{Santucci2016grail,Merrick2017value,Blaes2019}, and even boosting imitation learning techniques \cite{Nguyen2011}.

When facing the problem of autonomously learning multiple tasks, researchers typically focus on solutions where just one parametrised policy per task is sufficient to solve them \cite{Seepanomwan2017,Reinhart2017,Florensa2018}. Redundancy in motor control faced with autonomous agents is usually addressed as a problem and tackled through strategies such as goal-babbling \cite{Rolf2010,Rayyes2019}, which has been successfully implemented together with intrinsic motivation \cite{Baranes2013}. However, in complex environments presenting different contexts in time, the same task might need a set of different skills to be solved. 
The machine learning literature has proposed reinforcement learning models \cite{Sutton1998} to face problems where the transition and/or reward functions change in time and the agent is not informed on the context it is facing.
These type of problems have in particular been faced under the heading of ``non-stationary environments'' \cite{Padakandla2019} where the challenge is both to recognise the contexts to associate them with different models/policies and to manage all (possibly infinite) contexts \cite{daSilva2006,Nagabandi2018,Abdelfattah2019}.

Both in ``one-task/one-skill'' multi-task learning and in non-stationary setups, the issue of reducing the learning time has been addressed through \textit{transfer learning} techniques \cite{pan2009survey}: when learning a new policy, transferring previously acquired knowledge from a similar task can significantly speed up the process. Transfer learning has been profitably applied within the reinforcement learning framework \cite{taylor2009transfer}, however this technique does not cope with the problem of managing and reducing the generation of multiple contexts.
    
Differently from the research on non-stationary setups, in this work we consider the case where a robotic agent can use its sensors to perceive the contexts it is facing from time to time, and is allowed to tackle the same task using different strategies/policies in different contexts. A relevant challenge posed by these types of conditions is the possible ``explosion of contexts'': indeed, a real-world scenario presents possibly infinite set of features that might vary in time and generate a large number of contexts, thus overloading the agent's learning processes and memory resources. Moreover, since we are interested in the autonomous learning of multiple context-dependent tasks, the agent has also to manage the selection of the tasks not only with respect to its learning progress \cite{Santucci2013best} but also to the currently faced context.
To the best of our knowledge, this problem has not been previously faced with intrinsically-motivated autonomous learning agents. In the current work we thus compare various versions of the proposed architecture endowed or not with different mechanisms to face it the problem.

To face such types of problems, in this work we present Contextual-GRAIL (C-GRAIL), an extension of the GRAIL architecture (\textit{Goal-Discovering Robotic Architecture for Intrinsically-Motivated Learning} \cite{Santucci2016grail}). Similarly to its precursor, C-GRAIL is able to autonomously discover, select, and learn interesting tasks on the basis of intrinsic motivations, but new mechanisms allow C-GRAIL to assign tasks different values according to contextual features. Moreover, C-GRAIL is equipped with transfer learning capabilities and with a smart context-detector mechanism to cope with the problem of large numbers of possibly equivalent contexts.


\section{Problem description}
\label{sec:Problem}

\subsection{Autonomous learning of multiple tasks}
\label{sec:MultiTaskLearning}

From a reinforcement learning (RL) perspective, solving multiple tasks each one associated with a specific goal $g \in G$, defined in terms of a set of goal states $G \subset S$ can be seen as the objective of maximising different goal-dependent reward functions $r_g$ through an optimal policy described, as in \cite{Florensa2018}, as
\begin{equation}\label{eqn:GoalDependentPolicy}
   \pi^* = \underset{\pi}{\text{argmax}} \E_{g\sim P} \big[ R_g(\pi) \big],
\end{equation}
where $P$ is a probability distribution over the goals $G$, and $R
_g = \E[\sum_{t=0}^T r_{g,t}]$ is the expected return for the different tasks collected in the $T$ time steps of a trial.
The solution to the problem can be based on a set of policies $\pi_g(a|s)$, each associated to a goal $g$ and represented by a distinct module (or ``expert'') trained through any RL technique, or a single ``parametrised skill'' \cite{daSilva2014} defined as  $\pi(a|s,g)$.

In an autonomous open-ended learning perspective, we consider the case where the agent has to become competent to accomplish a possibly infinite set of tasks or goals. In particular, we assume that the agent's life is divided in two phases: a first phase (lasting a finite but unknown learning time $L$) of autonomous learning where the agent has to acquire the policies to achieve the different tasks, and a second phase where the agent is tested on a subset of the tasks drawn from an unknown distribution $P$ of tasks involving the same environment.

In the first phase, which is the focus of intrinsically motivated open-ended learning research and of this work, the agent has to maximise its \textit{expected competence} over the distribution of all possible goals $G$, where competence $C$ is a goal-dependent function quantifying the capability to obtain returns if tackling a particular task/goal, thus an approximation to the returns expected if $g$ is reinforced in the second phase. Equation 2 is a repetition of 1 and must be removed. In fact: $C_g = R_g$: it makes no sense to call the same quantity with two different symbols. Furthermore this implies that equation 2 is a repetition of equation 1. In this perspective, the objective function described in eq. \ref{eqn:GoalDependentPolicy} turns into
\begin{equation}\label{eqn:ComptenceObjFunc}
   \pi^*(a|s,g) = \underset{\pi}{\text{argmax}} \E_{g\sim P(g)} \big[ C(g) \big]
\end{equation}
\noindent
where $P(g)$ is the probability distribution of the goals, and where the set of policies $\pi^*_g$ in eq. \ref{eqn:GoalDependentPolicy} have been described as a parametrised policy $\pi^*(a|s,g)$.

Since the agent ignores the distribution $P$ of goals on which it will be tested in the second phase, during the autonomous learning phase the problem becomes the one of properly allocating the training time over all the possible goals (see also \cite{Santucci2019MGRAIL}). This can be tackled building a goal-selection policy $\Pi$ that learns to select the appropriate goal to practice at each trial of the first phase. Assuming that $C_t$ is the current expected competence of the system over the set of goals when using $\pi_t$, once a goal $g$ is selected the policy is trained for a certain amount of time $T$, resulting in a modification of the overall competence $C_{t+T}$. The aim of the agent is to select a sequence of goals for each trial so that after the entire learning period $L$ its competence over all the goals is maximised. The objective of the agent can thus be described as the construction of the goal-selection policy that identifies the goals to practice at each trial for a total of $L$ steps so that the resulting $C_L$ is maximal:
\begin{equation}\label{eqn:SelectionPolicyBandit}
   \Pi^* =  \underset{\Pi}{\arg\max} \,\, \E_{g\sim P} \big[ C_{t+L}(g) \,|\, \Pi \big].
\end{equation}

If we assume that tasks are independent (learning one task has no effect on the learning of other ones), the problem of task selection can then be modelled as an N-armed bandit \cite{Sutton1998} as in many architectures for autonomous open-ended learning \cite{Baranes2013,Santucci2016grail,Graves2017,Matiisen2019,Blaes2019}. Since it is not possible to analytically determine how the overall competence $C$ is changing after the selection and training of each goal, a common solution \cite{Santucci2013best} is to approximate eq. \ref{eqn:SelectionPolicyBandit} via a greedy approach that maximises the expected competence improvement $\Delta C$ defined as
\begin{equation}\label{eqn:CompImpr}
\Delta C_g = \E_{g\sim P(g)} \big[ C_{t+T}(g) - C_{t}(g) ]
\end{equation}
where $\Delta C^g$ is the expected competence improvement after practicing on one particular goal $g$ for $T$ steps. Note that given the transient nature of $\Delta C$, which changes with time and decreases while competence improves towards its maximum, the N-armed bandit associated with goal selection should then be considered as a \textit{rotting bandit} \cite{levine2017rotting}.

Under these assumptions, the general problem of autonomous learning multiple skills can thus be seen as a two-level problem: \textit{(a)} the high-level problem of selecting a goal $g$ to train on to rapidly maximise competence; and \textit{(b)} the low-level problem of improving the policies $\pi_g$ associated with the goals.

\subsection{Introducing context dependency}
\label{sec:ContextDependency}

As described in the introduction, usually multi-tasks learning is addressed assuming that for each goal $g$ a single policy $\pi_g$ might be sufficient to achieve the task. Differently, here we are considering the case where the same task might need different policies to be solved given different environmental conditions. This is a common situation in real-world scenarios where achieving the same goal may require a robot to use different strategies according to the current context. This may be due to the physical structure of the contexts: for example, the presence of certain objects in the environment could represent different obstacles to reach a certain target for an arm robot. The robot should hence possibly need to use different behaviours to reach for the same location in the different contexts. 

If we assume that the robot has the ability to identify the contexts on the basis of its sensors, we can divide the set of features $F$ perceived by the agent into two subsets: (1) ``low-level'' features $f_s$ changing at the fine time-scale of an attempt to achieve a goal and describing the state for $\pi_g$, for example the displacement of the joints of the actuator of the robot; (2) high-level features $f_{\phi}$ changing over a wider time-scale and describing the current context $\phi$ for $\Pi$, for example the position of the obstacles in the example mentioned above or more in general other features that are constant within the same context. Under these assumptions, the objective function related to goal-selection in eq. \ref{eqn:SelectionPolicyBandit} will then become

\begin{equation}\label{eqn:SelectionPolicyContextualBandit}
   \Pi^* =  \underset{\Pi}{\arg\max} \,\, \E_{g\sim P(g)} \big[ C_{t+L}(g) \,|\, \Pi (\phi) \big]
\end{equation}
\noindent so that goal-selection is now a contextual-bandit problem \cite{Sutton1998} in which the selection of the goal to train (and the attribution of values to the different goals) is dependent on the current context $\phi$. Similarly, the lower-level problem of training a skill for each goal is also dependent on the context: the system have to learn a set of policies $\pi_{g,\phi}$ to achieve the same goal in different contexts. This therefore involves that, in order to maximise the overall competence $C$, the system has to consider the specific competences for the goals with respect to the different contexts, $C_g(\phi)$, since competence improvement might be very different in different contexts.

This description of the problem under the assumption of context-dependent tasks assumes that each context $\phi$, described through its features $f_{\phi}$, is actually affecting goal achievability. However in real-world scenarios it is easy to imagine that many of the features generating a context \textit{do not} affect the policies that the system is trying to learn. The detected temperature, as well as if it is night or day, might have no effect on a reaching task performed indoor, while the on/off state of the light in the room could; similarly, the presence of objects distant from a target to reach should not be taken into consideration by the system, while objects positioned close to the target might become obstacles. If the artificial agent is supervised by a human designer, the latter could signal which elements of the context should be considered, but in an open-learning learning perspective the robot should autonomously manage the information from the sensors. 

Since everything \textit{could} be relevant, a simple solution would be to identify any distribution of the ``high-level'' features as a different context $\phi$, thus multiplying the number of contexts by the goals.
However, this strategy might substantially slow down the learning process and burden the system computational resources. Indeed, as illustrated in sec. \ref{sec:MultiTaskLearning}, the autonomous learning of multiple tasks is defined as a problem where the agent should maximise its overall competence in an unknown but finite time horizon. Transfer learning has proven to speed up the learning process (also with IMs \cite{uchibe2018cooperative}), but in a real-world open-ended scenario this might not be sufficient to cope with the combinatorial explosion of the contexts. Finding a smart and autonomous way to avoid context proliferation without impairing the learning of multiple, context-dependent tasks is thus a crucial issue for autonomous learning.


\section{Proposed solution}
\label{sec:Solution}

As illustrated in the previous section, developing artificial agents for the autonomous learning of multiple, context-dependent tasks presents two main challenges: (1) the allocation of training time over the tasks taking into account the different contexts; (2) the avoidance of context proliferation so as to not slow down the learning process under a finite and unknown time horizon.

To tackle the first issue, our approach follows the analysis in sec. \ref{sec:ContextDependency}: task selection is treated as an N-armed contextual-bandit where the system evaluates each goal/arm on the basis of the competence improvement $\Delta C$ expected from each goal in the current context $\phi$. More formally, the goal-selecting policy $\Pi(g|\phi)$ in eq. \ref{eqn:SelectionPolicyContextualBandit} at each moment selects a goal $g$ that maximises the immediate competence improvement given $\phi$, i.e.
\begin{equation}\label{eqn:ArgMaxContextual}
    \underset{g}{\arg\max} \, \, \Delta C (g|\phi)
\end{equation}
At the lower level of policy learning, for each goal and for each context the system associates a specific policy $\pi_{g,\phi}$ is trained through any learning algorithm maximising the reward $R_g$ related to goal $g$.

To tackle the second problem we propose to build on top of transfer learning techniques, such as those proposed in \cite{tommasino2016reinforcement}, to allow the transfer of knowledge between the policies $\pi_{g,\phi}$ achieving the same goal in different contexts. In particular, we introduce a heuristic based on the Ockham's razor:
the agent should tackle multiply contexts with different policies unless strictly necessary. Thus, instead of considering every new set of features $f_{\phi}$ as a different context, the agent starts to face it by actively ignoring the new contextual features, so that $\Pi$ can be seen as a contextual-bandit problem with just one ``baseline/blank'' context $\phi_0$. Only when ``something goes wrong'' the system considers the features to detect the  context. To asses the need to consider a new context as new with respect to the previously experienced ones, and within the model-free framework we use, the agent relies on the behaviour of its low-level motor policies $\pi_g$. In particular, novelty or surprise, as suggested by the intrinsic motivation framework \cite{Barto2013}, can be used to signal that there is an anomaly. Similarly to \cite{daSilva2014} (although this work was developed within a model-based framework), we suggest that an ``unexpected failure'' in achieving a goal can be used as a signal to identify new contexts (see sec. \ref{sec:CGRAIL_implementation} for details). In addition, having a fixed set of contexts for all the goals would be a violation of the Ockham's principle as a specific context could be relevant for certain goal but not for other goals. For this reason, the system has to identify a different subset of contexts $\Phi_g \subset \Phi$ for each goal, among all the possible contests $\Phi$, to further reduce context proliferation.

However, if this strategy is applied from very beginning of the learning process, when the agent is still performing (almost) random behaviours, this would still possibly result in an explosion of detected contexts. or this reason, our proposal is to trigger the context-detection mechanisms only when the competence $C_g(\sigma)$ for goal $g$ in the context $\sigma$ most similar to the current context $\phi$ is higher than a given threshold (the optimal setting of the threshold is beyond the scope of this work; see Appendix for the values used in this work), and the system detects a (negative) change in the reward function $R_g,$. Furthermore, considering all the contextual features of the environment would not follow the Ockham's principle. The system thus considers only those features in $f_{\phi}$ that are ``relevant'' for the task at hand: heuristics and previous knowledge might be used to this purpose, for example in a manipulation task they could be those related to a contacted object in proximity of the target. If these heuristics are not sufficient, the agent should also consider the other features.

Once identified, the relevant features will be recognised and considered by the system that will add a new context $\phi_{g, n+1}$ to the $n$ contexts already discovered for goal $g$, and such context will be considered by the contextual-bandit problem for goal-selection and a new policy $\pi_{g_,\phi_{g,n+1}}$ to achieve the same goal in the new context. Moreover, as previously stated, $\pi_{g,\phi_{g,n+1}}$ will possibly use transfer learning drawing knowledge from policies developed for the same goal in previously contexts (see sec. \ref{sec:CGRAIL_implementation} for an example).

\section{C-GRAIL}
\label{sec:C-GRAIL}

In this section, we describe Contextual-GRAIL (C-GRAIL), a system that extends the GRAIL architecture \cite{Santucci2016grail} and instantiate the proposals described in Sec. \ref{sec:Solution}. GRAIL is able to autonomously: discover goals; select goals according to competence-based intrinsic motivations (CB-IMs); recognise goal achievement; select computational resources to learn the goal-related skills. Despite its advancements, GRAIL (similarly to other autonomous open-ended learning systems \cite{Baranes2013,Blaes2019,Graves2017}) is not able to manage context-dependent task selection, nor to identify contexts and avoid their proliferation. Instead, C-GRAIL architecture can cope with multiple-context learning based on a ``smart context-detector'' mechanism that identifies and stores new contexts when needed, allowing the enrichment of the system behaviour. Although autonomous goal discovery is not the focus of this work, C-GRAIL is able to perform it inheriting this capability from GRAIL. 

Sec. \ref{sec:CGRAIL_general} presents a formal description of C-GRAIL capturing the main issues tackled in this paper.
Other non-relevant functions, such as goal-discovery and expert selection, together with the specific implementation of the system are reported in Sec. \ref{sec:CGRAIL_implementation} and in the Appendix.

\subsection{General functioning}
\label{sec:CGRAIL_general}

Algorithm \ref{algo:goalDisc} shows the general functioning of the architecture. We assume that the system operates within an episode-based reinforcement learning framework involving $trials$ each lasting $T$ time steps. We assume the state of the environment $s$ to be represented by two sets of different features: low-level features $f_s$ (such as proprioception) that might \textit{change within the trial}, and high-level contextual features $f_\phi$ (henceforth also denoted as $\phi$ for simplicity; these involve for example the presence of an obstacle in a certain location) that \textit{might change between the trials} and can be used to identify the context.
At the beginning of each $trial$, the system observes $s$ and passes $\phi$ to $CM$, a function returning the context $\phi_g$ actually perceived by each goal $g$: $\phi_g$ is obtained by filtering $\phi$ with a \textit{useful contextual feature} filter $ucf_g$ that allows the goal to ``see'' only the specific high-level features, among the features $\phi$, that in the past showed to be relevant for achieving the goal $g$. The filter $ucf_g$ is in particular formed by a list of the relevant high-level features. 

If a new context $\phi_{g,n+1} \not\in \Phi_g$ is identified for a goal, that context is added to the list of known contexts $\Phi_g$ of the goal and a new policy $\pi_{g,\phi_{g,n+1}}$ is created and associated to the goal. 
Transfer learning techniques are used to speed up the training of the policy (see sec. \ref{sec:CGRAIL_implementation} for details). To this purpose: a policy $\pi_{g,\phi_{g,k}}$ is randomly selected from the policies (if any) used to achieve the same goal in a different context $\phi_{g,k}$ and having a competence higher than a certain $transfer\_threshold$; the selected policy is used to act; if the policy successfully achieves the goal, then it is used to initialise the new policy.
For each trial this attempt to transfer is however done only with a certain probability so that the new policy with parameters created from scratch can progressively acquire the needed skill in case transfer is not possible.
%

The set of the identified current contexts $\phi_g$, one for each goal, are passed to the goal-selector policy $\Pi$ that determines, on the basis of competence improvement intrinsic motivations $\Delta C_{g,\phi_g}$, the goal to pursue in the current $trial$. We make no assumptions on how to implement $\Pi$, but it can be seen as a stochastic policy selecting one goal according to a $softmax$ distribution based on intrinsic motivations. The competence measure $C_g(\phi_g)$ is defined as the probability estimate of achieving $g$ given context $\phi_g$ and the system can autonomously asses it through a ``predictive function'' $\chi(g,\phi_g)$. $\Delta C_g(\phi_g)$ is thus computed as the difference between prior ($prior\_prob$) and posterior ($posterior\_prob$) estimated probabilities of achieving a goal given the same context $\phi_g$.

The selected goal $g$ and the identified context $\phi_g$ are then used to select the expert/policy to achieve $g$ (there can be just one expert for each couple $<g,\phi_g>$ or more than one, see sec. \ref{sec:CGRAIL_implementation}). 
At each time step $t$ of the trial, on the basis of the low-level features $f_s$ of the current state $s_t$, an action $a_t$ is selected through the policy $\pi_{g,\phi_g}(f_s)$ bringing the agent in $s_{t+1}$. The new state is observed and a reward is autonomously calculated using the $GM$ function: this function returns $1$ (\textit{success}) if the system has achieved the goal-state associated with $g$, and $0$ (\textit{failure}) otherwise.
The policy $\pi_{g,\phi_g}$ is hence updated accordingly using such reward. If the $trial$ ends before its time-up $T$ with a $failure$ (see sec. \ref{sec:CGRAIL_implementation}), and the computed $prior\_prob$ is over a certain $competence\_threshold$, the high-level features $f_{fail}$ of $f_{\phi}$ related to the condition at hand (for the specific heuristics used here see sec. \ref{sec:CGRAIL_implementation}) are added to the list of $ucf_g$. Finally, the competence predictor $\chi$ is updated accordingly to goal achievement $[0;1]$ and the difference between $prior\_prob$ and $posterior\_prob$ is calculated to provide the CB-IM $\Delta C_g(\phi_g)$ that biases goal selection.

\begin{algorithm}
Let $known\_goals = \{\}$ be the list of discovered goals  \\
\ForEach {$g \in known\_goals$} {
$ucf_g = \{\}$ \algocomment{identified contextual features $f_{\phi}$}\;
$\Phi_g = \{\}$ \algocomment{identified contexts $\phi_g$} \;
}
Let $\Delta C_g(\phi_g)$ be the agent's intrinsic motivation for achieving goal $g$ in context $\phi_g \in \Phi_g$\; 
\vspace{0.2cm}
\For{$\mbox{trial} \gets 1$ \textbf{to} $\ldots$} {
    Set the initial configuration of the environment $s_0$\;
    Let $t$ be the current time step within the trial\;
    \vspace{0.2cm}
    \algocommentblock{Determine current context for each goal, and store new context and policy}\;
    Observe current state of contextual features $\phi$\;
    \ForEach{$g \in known\_goals$}{$\phi_{g,n+1} = CM(f_{\phi}, ucf_g)$\;

    \If{$\phi_{g,n+1} \not\in \Phi_g$}{
        $\Phi_g \leftarrow \Phi_g \, \cup \, \phi_{g,n+1}$ 
        \algocomment{add new context}\;
        Create new policy $\pi_{g,\phi_{g,n+1}}$\;
          With a certain probability use the new policy parameters, otherwise randomly select a policy $\pi_{g,\phi{g,k}}$ among those, if any, with a $competence \geq transfer\_threshold$\;
          $\pi_{g,\phi_{g,n+1}}\xleftarrow{transfer}\pi_{g,\phi_{g,k}}$\;
        }
    }
    
    \algocommentblock{Select a goal and related policy}\;
    Select a goal $g$ based on $g \sim \Pi(\phi_g, \Delta C_{g,\phi_g})$
    \algocomment{select goal given identified contexts and goal intrinsic motivation}\;
    $\mbox{prior\_prob} \leftarrow \chi(g | \phi_g)$\algocomment{probability of achieving $g$ given $\phi_g$}\;    
        \vspace{0.2cm}
        \For{each time step $t \gets 1$ \textbf{to} $T$ within the trial} {
            Select an $a_t \sim \pi_{g,\phi_g}(f_s) $\;
            Perform action $a_t$ in the environment\;
            Observe next state $s_{t+1}$\;
            Let $r_t = GM(s_{t+1}, g)$\;
            Update $\pi_{g,\phi_g}$ \algocomment{update policy based on $(s_{t}, a_t, r_t, s_{t+1})$ }\;
            \If{failure \& $t<T$ \& $prior\_prob>competence\_threshold$} {
                Observe features $f_\phi$ related to $s_{t+1}$\;
                Let $f_{fail}$ be failure-related features in $f_\phi$\;
                \If{$f_{fail} \not\in ucf_g$}{$ucf_g \leftarrow ucf_g \, \cup \, f_{fail}$}
                
            }
            
        }   
        \vspace{0.2cm}
        \algocommentblock{Update success predictor and intrinsic motivation}\;
        Update goal-success predictor $\chi$ based on $GM$\;
        $\mbox{posterior\_prob} \leftarrow \chi(g | \phi_g)$ \algocomment{updated probability of achieving goal}\;
        $\Delta C_g(\phi_g) \gets \mbox{posterior\_prob} - \mbox{prior\_prob}$\; 
}
\caption{C-GRAIL: contexts handling}
\label{algo:goalDisc}
\end{algorithm}

\subsection{Implementation}
\label{sec:CGRAIL_implementation}

\begin{figure}
    \includegraphics[width=\linewidth]{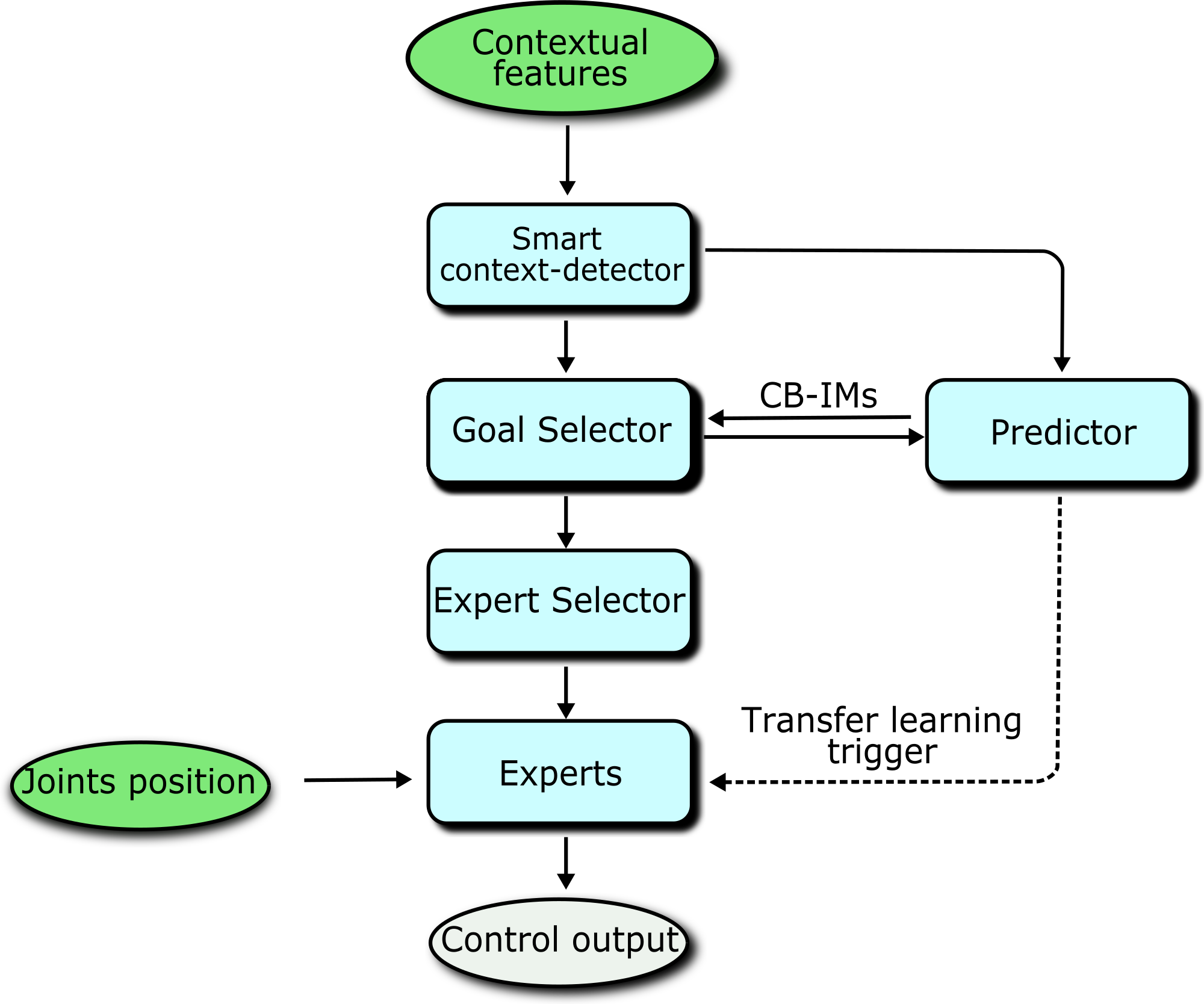}
    \caption{A sketch of C-GRAIL architecture, with main components and connections. Note that experts and goal and context specific.}
    \label{fig:CGRAIL_Arch}
\end{figure}

This section presents the implementation of C-GRAIL in the current work as sketched in Fig. \ref{fig:CGRAIL_Arch}. Further technical details can be found in the Appendix.

\subsubsection{The robot and the sensory input}
\label{sec:robotAndInput}

C-GRAIL is implemented in a simulated iCub robot through the 3-D physical engine GAZEBO and dedicated YARP plugins. We use two arms of the robot with 4 degrees-of-freedom, since wrists joints are kept fixed and hands are substituted by 2 scoops (see Fig. \ref{fig:environment}). Collisions are not taken into consideration: a sensor at the centre of each scoop detects whether the robot has ``touched'' one of the objects in the environment. The input to the robot are provided by two sources: the camera of its right eye, kept fixed so that all objects remain in the visual field; the arms joints whose angles determine the robot proprioception.

\subsubsection{Goal discovery}
\label{sec:goalDiscovery}

As in other versions of the GRAIL architecture \cite{Santucci2016grail,Santucci2019MGRAIL}, the system is endowed with a mechanism to autonomously discover ``interesting'' states and store them as possible goals: here we use a simple, biologically inspired \cite{jacquey2019sensorimotor,Sperati2018} built-in strategy detecting those states resulting from a change in the visual input. 
The visual input images corresponding to two succeeding time steps are compared and whenever there is a change (the movements of the arms are excluded, see Appendix) the image resulting from the difference between the two images is taken and compared with previously stored ones: if new, it is stored in the goal-representation map (GR-M) and associated with the first available unit of the goal-selector (see sec. \ref{sec:goalselector}). Since the system retains the representations of the discovered goals, it is able to autonomously assess, through a goal-matching function (GM, sec. \ref{sec:goalmatching}), if a goal has been achieved, thus providing a signal to train the policies (the experts), the expert-selector, and the predictor determining the intrinsic motivations (sec. \ref{sec:IntrinsicMotivations}).

\subsubsection{Smart context-detector}
\label{sec:ContextDetection}

This is the component introduced in C-GRAIL to avoid context proliferation. It is composed of two mechanisms: the detection of current relevant contexts and the gathering of \textit{useful contextual features} ($ucf$). At the beginning of each trial, the smart context-detector (SCD) receives as input the set of contextual feature $\phi$ in the environment. These are the values of those features that are not changing during the trial, which in the experimental case presented here (see sec. \ref{sec:ExperimentalScenario}) are the occupied/non-occupied possible positions of the obstacles. For each goal $g$ the system has a list of `identified' $ucf_g$ that have this far revealed relevant to accomplish the goal. A \textit{Context Matching} function $CM$ filters the current features $\phi$ with the $ucf_g$ of each goal, returning a list of goal-specific contexts $\{\phi_{g_1}, \phi_{g_2}, \ldots, \phi_{g_n}\}$ for all the currently \textit{known goals}. This list is then used by the goal-selector to choose the goal to pursue on the basis of context-specific intrinsic motivations. If a new context $\phi_{g,n+1}$ is detected for goal $g$, it is added to the set of $known contexts$ $\Phi_g$ for that goal, and a new policy $\pi_{g,\phi_{g,n+1}}$ associated to the goal $g$ is added to the repertoire of the system as a new expert (see sec. \ref{sec:Experts} and \ref{sec:TransferLearning}).

While training on $g$, if there is a $failure$ that terminates the trial before its maximum duration $T$ (see sec. \ref{sec:ExperimentalScenario}), and if the predicted performance level (see sec. \ref{sec:IntrinsicMotivations}) within the current context $\phi_g$ is higher then a $competence\_threshold$ (whose value has been set through experimental heuristics, see Appendix for details), the SCD component receives in input the values of the contextual features $\phi$ related to the environmental state $s_t$ (here represented by the centre of the scoop on the actuated arm) where the agent is situated when the trial terminates. In other words, if the agent bumps into an obstacle, only the features related to that specific obstacle are passed to the SCD, while all the others are not taken into consideration since the system, following the law of parsimony, assumes that only the features of its current position are involved in the signalled failure. These $\phi$ are then compared with the previously identified list $ufc_g$ for the pursued $g$ and if not present they are added to it.

\subsubsection{Goal-selector}
\label{sec:goalselector}
At each trial, the goal-selector determines the goal to pursue on the basis of CB-IMs and the current context. In particular, the goal-selector takes as input the current goal-specific contexts $\phi_g$ identified by the SCD component and selects the goal to pursue according to a \textit{softmax} selection rule based on the current values of the goals, updated through a standard exponential moving average (EMA) based on goal-and-context specific CB-IMs (sec. \ref{sec:IntrinsicMotivations}).
As in other GRAIL versions, at the beginning of the experiment the goal-selector is a ``blank vector'' whose units are not associated with any specific goal but during exploration the system is able to autonomously discover new goals and associate them to the units in the goal-selector as described in sec. \ref{sec:goalDiscovery}.

\subsubsection{Expert selector}
\label{sec:ExpertSelector}

In its simplest version, C-GRAIL can be implemented associating to each $<g,\phi_g>$ couple a specific expert encoding the policy that controls the robot when learning $g$ in $\phi_g$: when a task is selected through the goal-selector, the expert associated with that goal-context couple gets in charge of guiding the robot behaviour. However, as in previous versions of GRAIL, we endow the robot with the capability of autonomously selecting different computational resources, one for each arm, to achieve the same goal even in the same context. All the goals (within each context) can be achieved using both arms, but in this way the system has a further degree of autonomy whose advantages, investigated in \cite{santucci2014autonomous}, are beyond the scope of this work.
Given the selected goal and the current context for that goal, the selection of the expert is based on an EMA of the rewards provided by the goal-matching (GM) function for achieving that goal. Based on this mechanism, the higher the probability of success with an arm, the higher the probability of selecting it for the same goal in the same context.

\subsubsection{Experts}
\label{sec:Experts}
Each expert is implemented as a neural-network actor-critic model modified to work with continuous states and action spaces \cite{Doya2000}. When selected, each expert receives as input the low-level features $f_s$ corresponding to the four actuated joints of the related arm (three for the shoulder, one for the elbow). Four output units of the expert's actor encode the displacement of the joints through position control. At each step, the selected expert is trained through a TD reinforcement learning algorithm \cite{Sutton1998} maximising the rewards generated by the GM function for achieving the currently selected goal.

\subsubsection{Transfer learning}
\label{sec:TransferLearning}
After a goal has been selected, the system checks through its competence predictor $\chi$ (see sec. \ref{sec:IntrinsicMotivations}) the expected performance for that goal in the current context $\phi_{g,k}$ for the associated policy $\pi_{g,\phi_{g,k}}$. If the prediction is under a certain \textit{learning threshold} (see Appendix for all the details), the system checks if a policy $\pi_{g,\phi_{g,j}}$ exists for the same goal in a different context $\phi_{g,j}$ whose expected performance is higher than a \textit{transfer threshold} and, with a certain \textit{transfer probability}, uses it for transfer learning (if multiple candidate experts are available, one is randomly picked with uniform probability); the robot action is controlled by the policy of the selected \textit{source expert}
and if the trial ends achieving the current goal the parameters of the policy (actor) and evaluation function (critic) of the \textit{source expert} are copied into the \textit{target expert} associated with the goal at hand. Instead, in the case of failure the transfer does not happen and is possibly attempted again in the next trial with a different (or the same) expert.

\subsubsection{Goal-Matching}
\label{sec:goalmatching}
The GM function allows the agent to autonomously check if a pursued task has been achieved. When a task is selected by the goal-selector its representation in the GR-M is activated. While operating, if a change is detected in the visual input, GM compares it with the representation of the currently selected goal: the component generates a signal of $1$ or $0$ if there is or there is not, respectively, an overlap of the two images (see Appendix for details). In the specific experimental domain presented here the GM generates a binary signal, but this signal could be continuous in different domains \cite{Seepanomwan2017}. The signal is used as a reward signal to train the experts and the expert-selector; moreover, it is used as a teaching input for the predictor whose activity determines the CB-IM signals that bias goal selection.

\subsubsection{Intrinsic Motivations}
\label{sec:IntrinsicMotivations}

The CB-IMs signal, modelled as a competence improvement signal \cite{Santucci2013best}, is the result of the activity of the competence predictor $\chi$ . At the beginning of each trial, this component receives as input the selected goal $g$ and the goal-specific context $\phi_g$, and outputs the predicted performance of the agent, which is used also in the transfer learning process (sec. \ref{sec:TransferLearning}).
At the end of the trial the prediction is updated according to the GM output, and the competence prediction improvement (CPI), calculated over a fixed time-window (see Appendix), determines goal/context-dependent CB-IM signals biasing goal selection.

\section{Experimental setup}
\label{sec:ExperimentalScenario}

\subsection{Environment and task}
\label{sec:Environment}

\begin{figure}[htb]
    \centering
    \includegraphics[width=\linewidth, 
    height=0.7\linewidth]{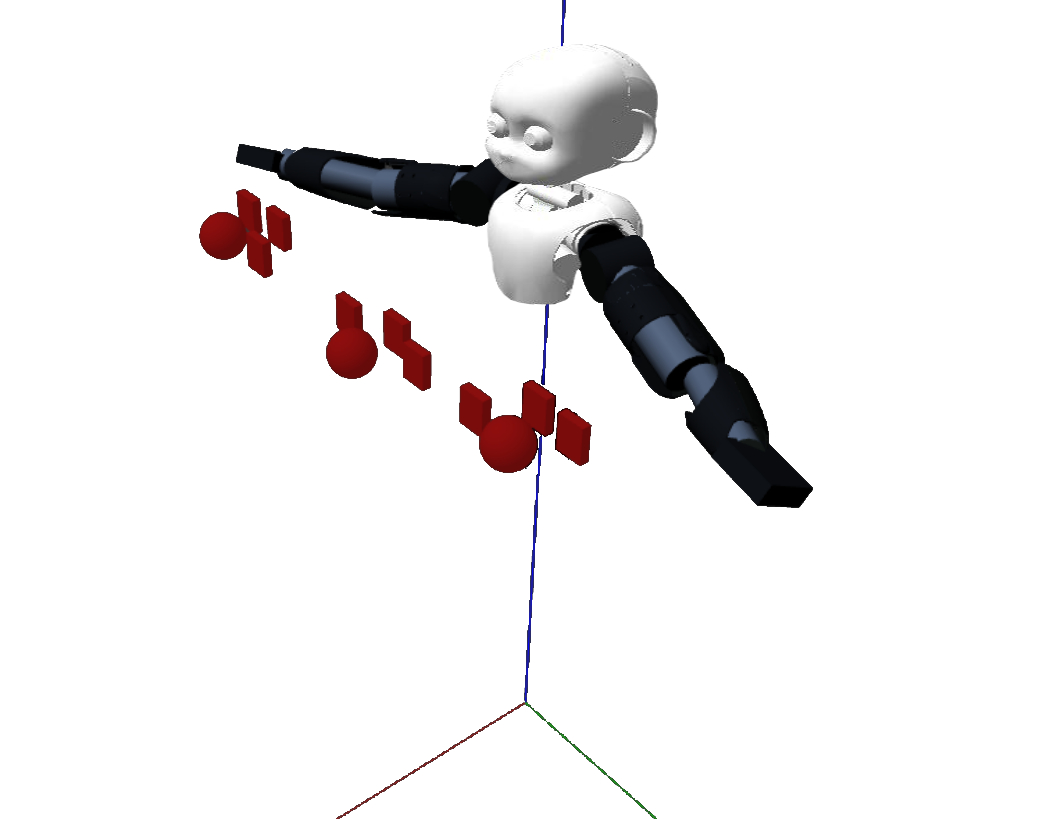}
    \caption{Setup used to test the systems. The figure shows the robot, of which we used here only the two actuated arms and a camera in the head, the three target spheres, and the nine obstacles.}
    \label{fig:environment}
\end{figure}

In addition to the simulated robot, the environment is composed of 3 spherical objects representing potential reaching targets (Fig. \ref{fig:environment}).
Moreover, a varying number of rectangular parallelepiped obstacles (maximum 9) are placed close to the targets. In particular, for each target there might be simultaneously 3 obstacles facing it on its right, left, and middle. All the objects are anchored to the world and they are always in the visual field of the robot and at reaching distance. When touched, a target `switches on' by changing its colour to green. The task consists in learning to activate the targets in the shortest amount of time and in the different environmental contexts. At the beginning of each \textit{trial}, the number of the obstacles change thus configuring a new context (with 9 present/absent obstacles there are $2^9=512$ possible contexts). The experiment lasts for 50,000 trials each ending after 700 steps or when one any object (targets or spheres) is touched. 

\subsection{Compared systems}
\label{sec:ComparedSystems}

To test the efficacy of our approach, we compared \textbf{C-GRAIL} to other three systems, each one lacking some of the mechanisms implemented in our architecture and reflecting other systems in the literature:

\begin{itemize}
    \item \textbf{Bandit: multi-armed bandit.} This system is built similarly to other architectures for autonomous multi-task learning \cite{Santucci2016grail,Blaes2019,Baranes2013}, where goal selection is treated as a multi-armed bandit without context and each task/goal is associated with a single learning policy. No transfer learning (TL) is used.
    \item \textbf{C-Transfer: contextual bandit with transfer learning.} This system implements autonomous goal-selection as a contextual bandit similarly to what done in \cite{Forestier2017,Oddi2020}, but it is not endowed with the smart context-detector (SCD) mechanism as C-GRAIL, thus all the possible contexts are actually taken into consideration by the system. The system is equipped with TL to share knowledge between policies achieving the same goal in different contexts.
    \item \textbf{Smart C-Bandit: contextual bandit with smart context-detection but no TL.} This system implements the SCD mechanisms as C-GRAIL but it lacks TL. The system is thus able to limit context proliferation but it has to learn each context-dependent expert from scratch.
\end{itemize}

\section{Results}
\label{sec:Results}

In this section we present the results of the caparison of the four systems: Bandit, C-Transfer, Smart C-Bandit, C-GRAIL.

\begin{figure*}
    \centering
    \begin{subfigure}[t]{0.49\textwidth}
    \includegraphics[width=\linewidth, height=0.7\linewidth]{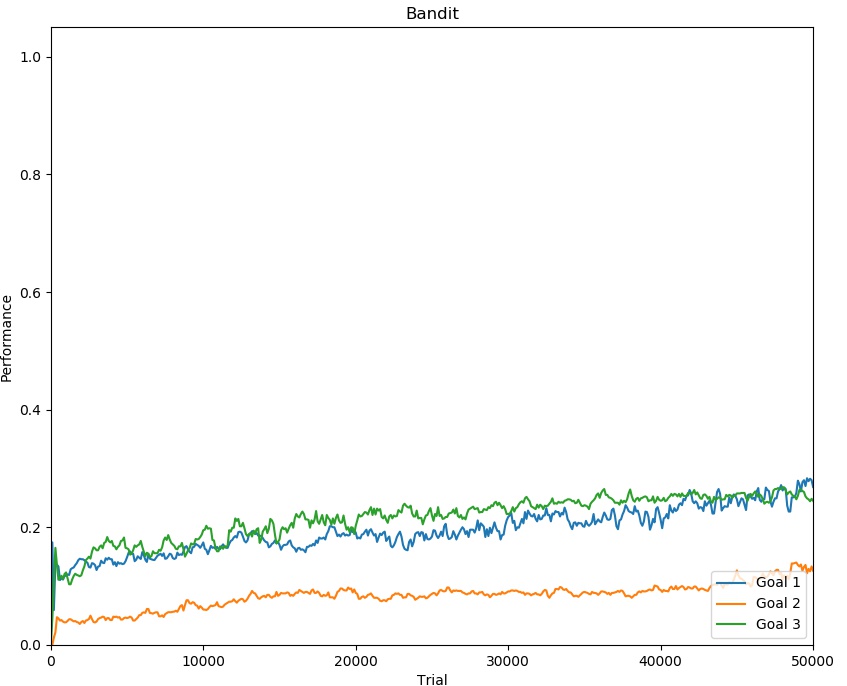}
    \caption{Bandit}
    \label{fig:BanditGen}
    \end{subfigure}
    \hfill
    \begin{subfigure}[t]{0.49\textwidth}
    \includegraphics[width=\linewidth, height=0.7\linewidth]{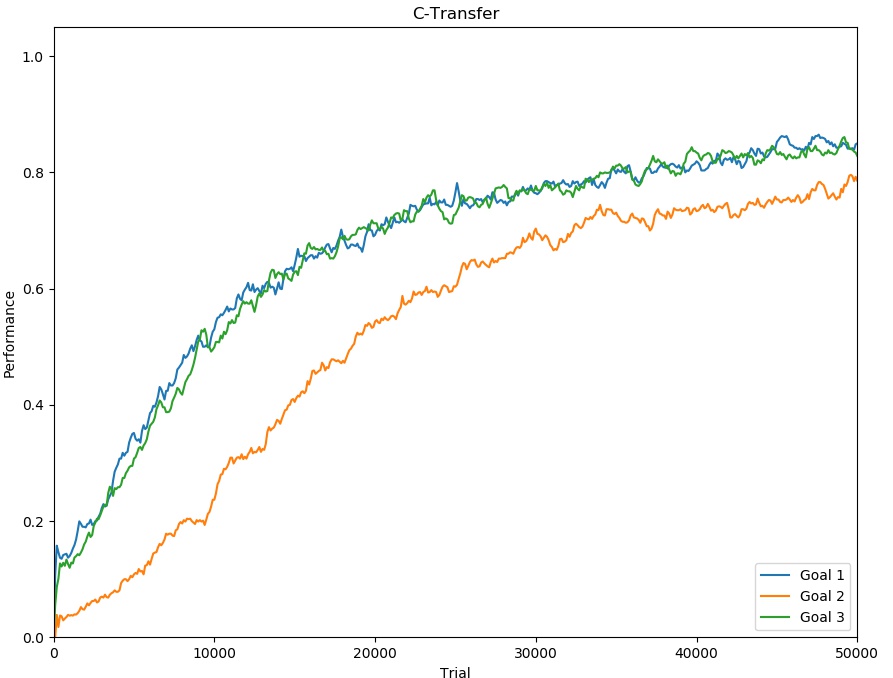}
    \caption{C-Transfer}
    \label{fig:CTransGen}
    \end{subfigure}
    \medskip
    \begin{subfigure}[b]{0.49\textwidth}
    \includegraphics[width=\textwidth, height=0.7\linewidth]{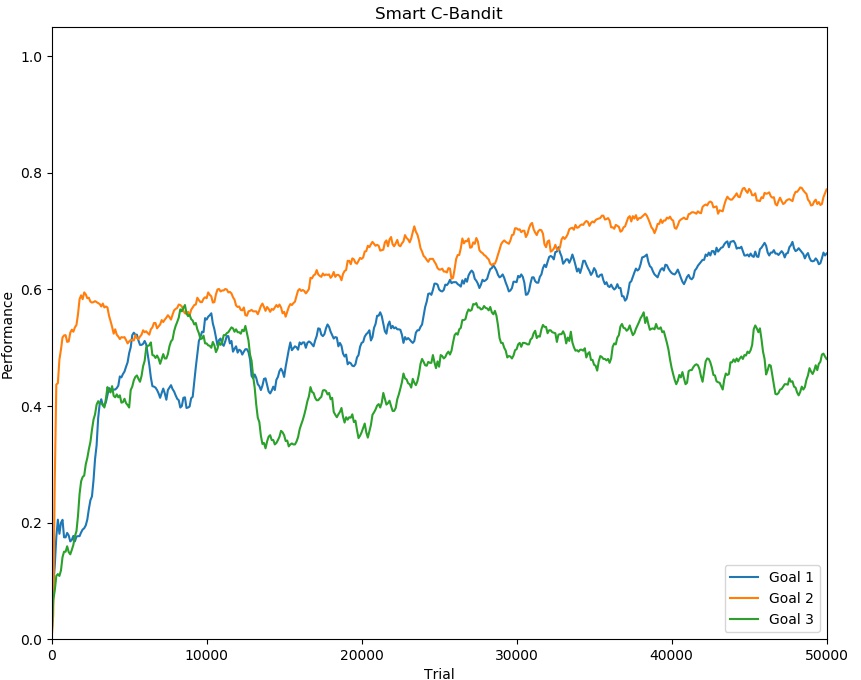}
    \caption{Smart C-Bandit}
    \label{fig:SmartCGen}
    \end{subfigure}
    \hfill
    \begin{subfigure}[b]{0.49\textwidth}
    \includegraphics[width=\textwidth, height=0.7\linewidth]{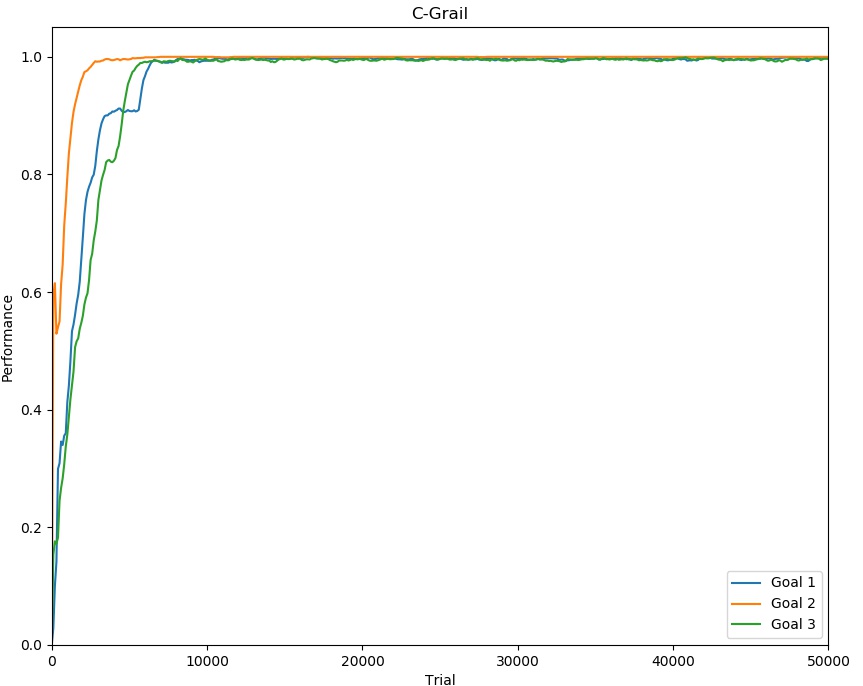}
    \caption{C-GRAIL}
    \label{fig:CGrailGen}
    \end{subfigure}
    \caption{The average performance of the four tested systems in learning the 3 tasks}
    \label{fig:GeneralPerformance}
\end{figure*}

Fig. \ref{fig:GeneralPerformance} shows the average performances (over 10 repetitions) of the four systems engaged in learning to reach the three different spheres. As expected, the Bandit cannot achieve a sufficient performance on the tasks. The system is not able to distinguish the different contexts and has only one trainable expert for each goal: not only the agent cannot select the most profitable goals according to the current obstacles, but having to learn a single policy to reach the same goal in different contexts it will continue to modify the same experts thus incurring catastrophic forgetting.

C-Transfer achieves a good performance over all the tasks ($\sim$80\%) only at the very end of the experiment.
Smart C-Bandit reaches a lower average performance (between $\sim$75\% on goal 2 and $\sim$45\% on goal 3).
C-GRAIL is capable to achieve a 100\% performance after only $\sim$8,000 trials. To better investigate the differences between these last three systems and to understand how the components of C-GRAIL contribute to its performance, we focus the analysis on the learning of a single goal (goal 1, i.e. the sphere positioned on the left of the robot) considering data of a single representative seed for each system.

\begin{figure}
    \centering
    \includegraphics[width=\linewidth, height=0.7\linewidth]{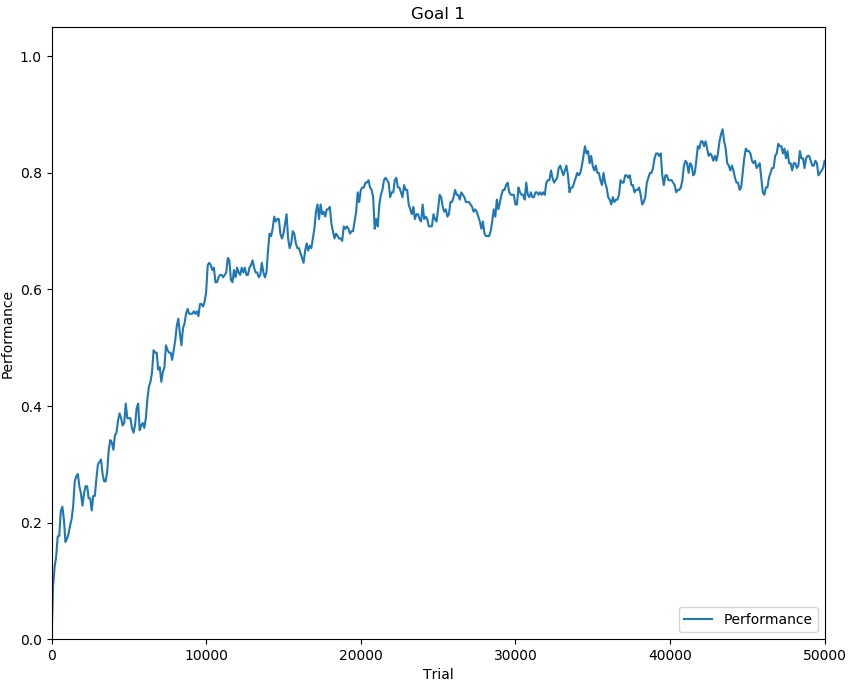}
    \caption{C-Transfer: performance on goal 1.}
    \label{fig:CTransGoalOne}
\end{figure}

\begin{figure}
    \includegraphics[width=\linewidth, height=0.7\linewidth]{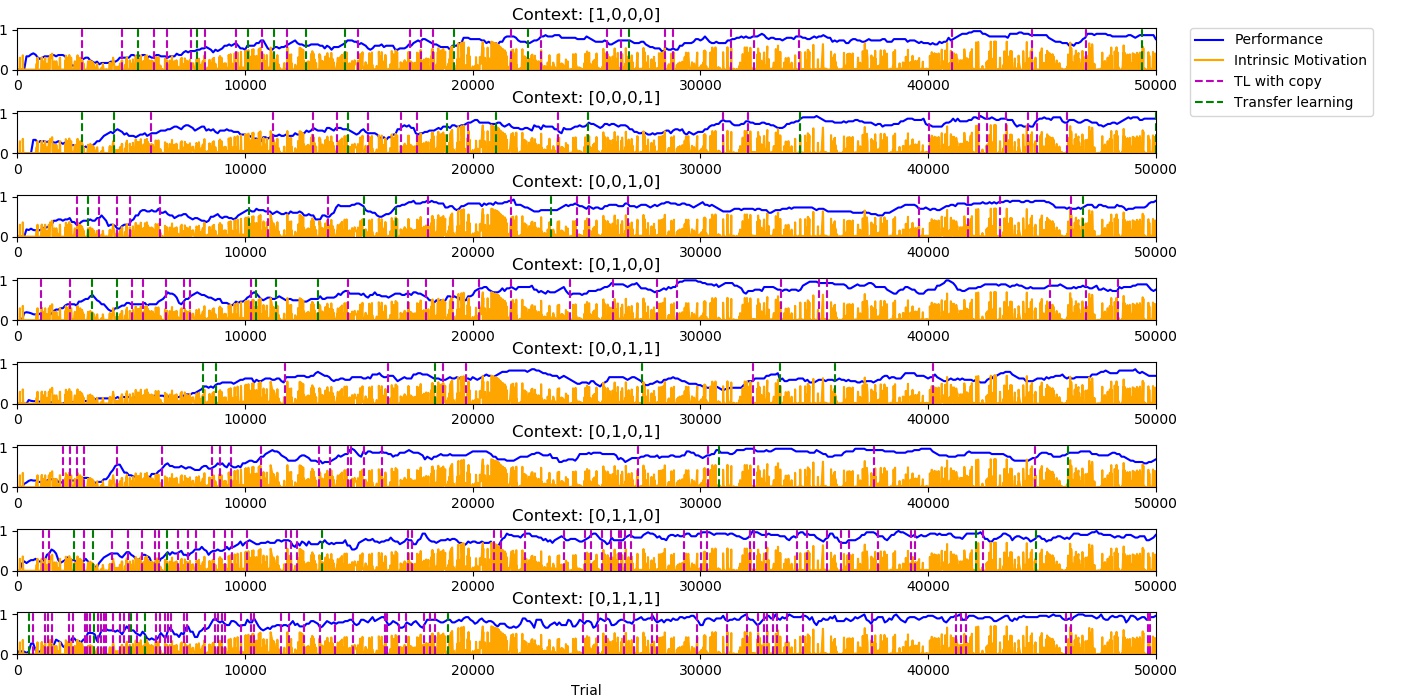}
    \caption{C-Transfer: analysis of context-related learning for goal 1, with transfer learning between contexts and intrinsic motivation signal}.
    \label{fig:CTransContexts}
\end{figure}

Fig. \ref{fig:CTransGoalOne} shows the performance of C-Transfer on goal 1 as an average over all the contexts, while Fig. \ref{fig:CTransContexts} shows the learning and performance for that goal with respect to the different contexts depending on the configuration of the obstacles. To make the visualisation easier, we have shown here only the contexts determined by the three obstacles close to the sphere associated with goal 1, i.e. $2^3$ contexts: the one without obstacles (identified as 1,0,0,0 in Fig. \ref{fig:CTransContexts}) and the various combinations of the three. However notice that the total number of possible contexts is given by the combination of the nine obstacles that may be present in the world ($512$ possible configurations). Since C-Transfer does not have the smart context-detector embedded both in Smart C-Bandit and C-GRAIL, the system considers, for each goal, all the 512 possible contexts, thus 512 policies to potentially train. However, the transfer mechanism guarantees the system to share the motor competence acquired in the contexts, both with partial transfer (dotted green lines in Fig. \ref{fig:CTransContexts}) or copying the entire policy of a different context when it can achieve the goal (dotted purple lines in Fig. \ref{fig:CTransContexts}). Although C-Transfer manages to achieve a high performance, the learning process is slowed down by the fact that the system considers all the 512 contexts, including those combined with obstacles not relevant to the task at hand (i.e. those close to the other spheres). This not only wastes time in transferring skills between contexts, but also slows down the goal selection process: as shown in the fig. \ref{fig:CTransContexts}, the intrinsic motivations signal (orange lines) is always present on each goal-related context even when competence has been properly acquired. This is because the plotted signals incorporate the combination of the single goal-related contexts with all the configuration of the obstacles related to the other spheres. Since intrinsic motivations are context-related, the system might still have motivation to improve its competence in a context even if it is actually able to achieve the goal (if we just consider the configuration of the three goal-related obstacles): this is clear if we look at the final trials of the simulation, where even if the performance is near the 80\% the magnitude of the IM signal is high, and where the system is still performing transfer learning between contexts (which are those not explicitly reported in the graph).

\begin{figure}
    \centering
    \includegraphics[width=\linewidth, height=0.7\linewidth]{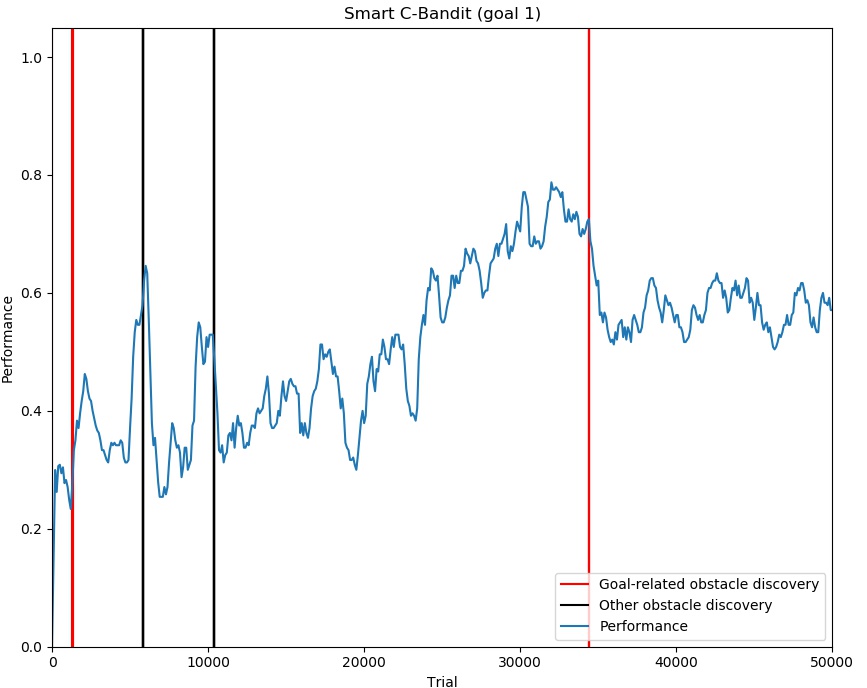}
    \caption{Smart C-Bandit: performance on goal 1 and discovery of contexts.}
    \label{fig:SmartCGoalOne}
\end{figure}

\begin{figure}
    \includegraphics[width=\linewidth, height=0.7\linewidth]{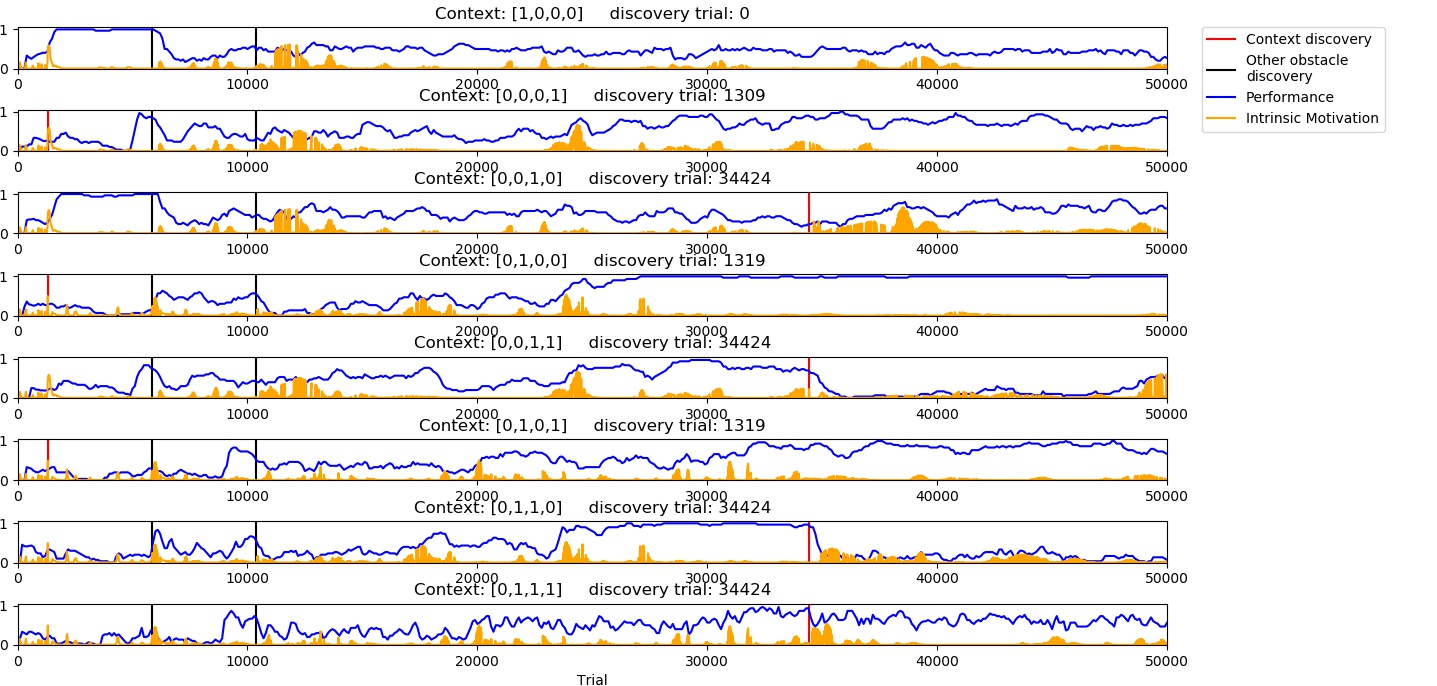}
    \caption{Smart C-Bandit: analysis of context-related learning for goal 1, with context discovery and intrinsic motivation signal.}
    \label{fig:SmartCContexts}
\end{figure}

Fig. \ref{fig:SmartCGoalOne} shows the performance of Smart C-Bandit system on goal 1, averaged on all the possible contexts. Thanks to the smart context-detector described in Sec. \ref{sec:ContextDetection} the system adds new context-related policies only when hitting on obstacles (and only if the current policy has reached a minimum level of performance). In particular, when performing task 1 Smart C-Bandit ``discovers'' two obstacles close to the associated sphere (dotted red lines in Fig. \ref{fig:SmartCGoalOne}) and only two other obstacles standing close to the other targets (dotted black lines in Fig. \ref{fig:SmartCGoalOne}), resulting in four goal-related contexts plus all the combinations with the other two discovered obstacles for a total of ``only'' 16 contexts, way less than the 512 that a system such as C-Transfer has to consider. However, this advantage is not enough to make Smart C-Bandit perform better than C-Transfer. Indeed, not being able to exploit the power of transfer learning, every time a new context is identified the system has to start learning the related skill from scratch, thus loosing time in re-acquiring previously learnt behaviours. This is clear from Fig. \ref{fig:SmartCContexts}: in many cases, when the system discovers a new obstacle/context there is a significant drop in the performance even when the agent was properly accomplishing the task. 

\begin{figure}
\includegraphics[width=\linewidth, height=0.7\linewidth]{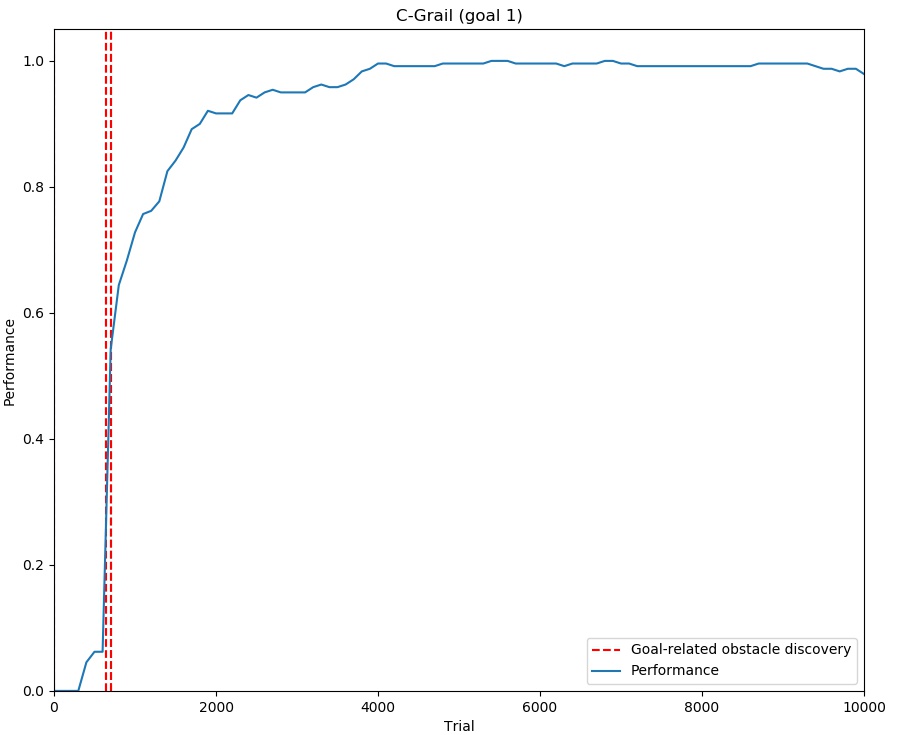}\llap{\raisebox{1cm}{\includegraphics[height=4.5cm]{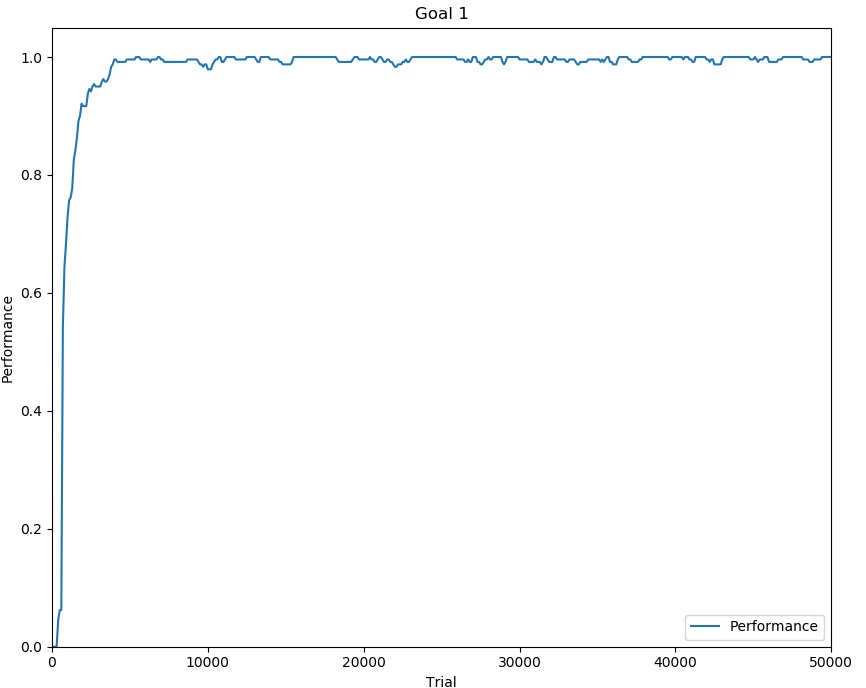}}}
\caption{C-GRAIL performance on goal 1, averaged over all the contexts. Small figure: entire simulation, 50,000 trials. Big figure: zoom on 10,000 trials and timing of the new context discoveries.}
\label{fig:CGRAILPerfGoal}
\end{figure}

If transfer learning alone is not able to reduce learning time due to the large number of contexts generated by only nine obstacles, and if the smart reduction of context alone cannot cope with the problem of having to re-learn skills from scratch, integrating the two mechanisms into one architecture results in a profitable solution to the autonomous learning of multiple context-dependent tasks. Fig. \ref{fig:CGRAILPerfGoal} shows the performance of C-GRAIL on goal 1 averaged over the different contexts (for a video of the behaviour of the robot, see: \url{https://www.youtube.com/watch?v=33EnmAbfXSo}; the video shows an initial phase where the robot explores randomly, an intermediate phase where the robot has started to learn, and a final phase where+ the robot shows a highly efficient behaviour). The system reaches a 100\% performance at trial $\sim$4,000 and discovers two task-related obstacles as Smart C-Bandit but no other obstacles, thus employing only four different contexts for this task. Thanks to transfer learning, the system has no drop in performance when new contexts are identified and a rapid learning of the four experts associated with the four identified contexts, and can thus easily cope with all the 512 contexts present in the environment. This guarantees fast learning, reflected on a rapid decrease in the IM signals related to goal 1, which allow the system to focus and fast learn also the other tasks (as reported in Fig. \ref{fig:CGrailGen}). Moreover, the fast learning also helps the system to reduce the amount of discovered obstacles: the quicker the system learns the task, the less exploration and therefore the lower the risk of hitting and discovering other obstacles.

\begin{figure}
    \includegraphics[width=\linewidth, height=0.7\linewidth]{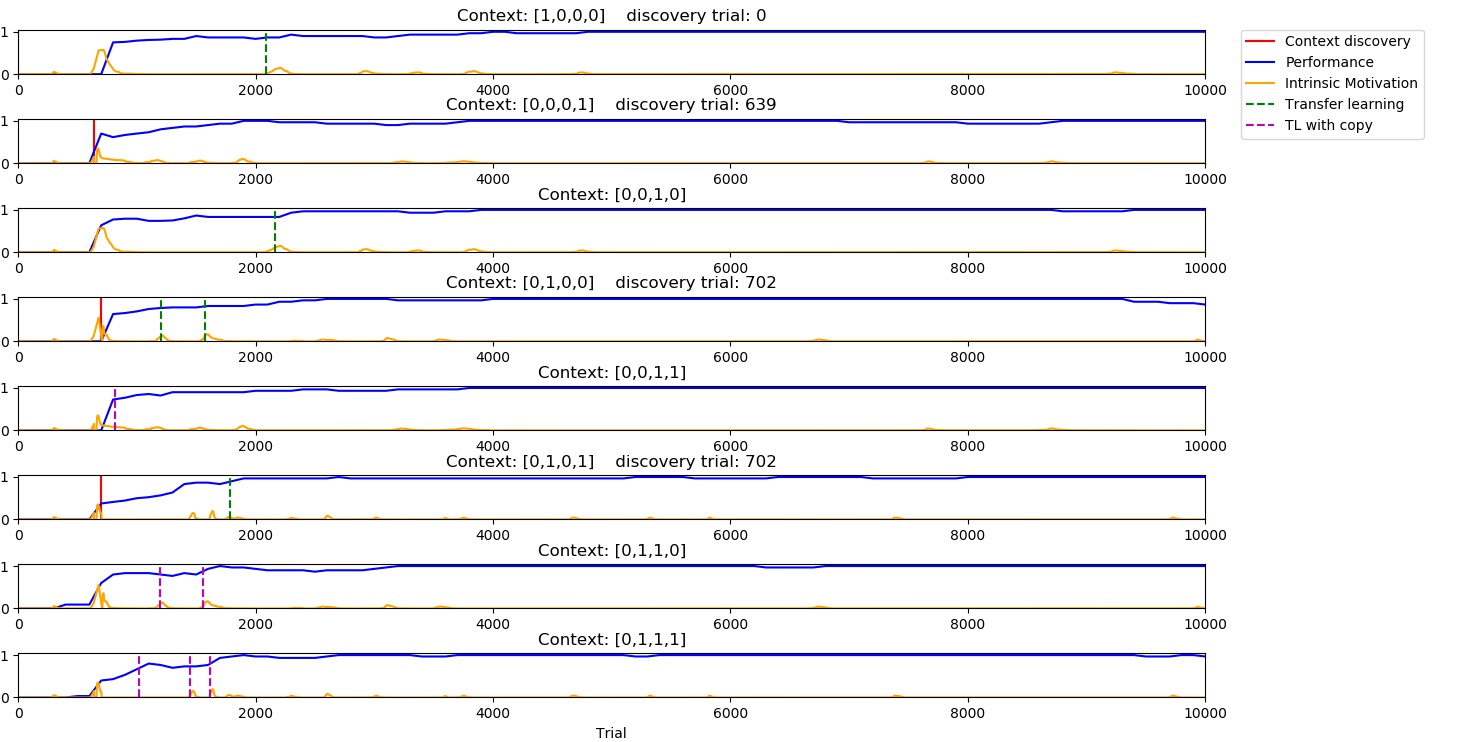}
    \caption{C-GRAIL: analysis of context-related learning for goal 1 on the first 10,000 trials, with context discovery, transfer learning between contexts, and intrinsic motivation signal}.
    \label{fig:SmartCContexts}
\end{figure}

\section{Discussion and conclusions}
\label{sec:Conclusions}
In this work we tackled the problem of the autonomous learning of multiple context-dependent tasks. What we proposed is to combine the well known practice of transfer learning with a mechanism that guarantees the reduction of the number of contexts to handle, focusing only on those that are relevant for the tasks to accomplish. To this purpose, we presented C-GRAIL, an integrated robotic architecture for intrinsically motivated open-ended learning that enhances the previous GRAIL system \cite{Santucci2016grail} with mechanisms for autonomous multiple task learning, smart context-detection, and transfer learning. We compared C-GRAIL with other systems in a simulated robotic scenario and we shew how its components contribute to the rapid and successful learning of multiple, context-dependent tasks.

In particular, the results show that transfer learning alone ensures the rapid sharing of acquired skills, but it cannot be sufficient in environments with multiple, possibly not relevant contexts. On the contrary, a system able to take into account only the conditions that actually influence the success of the tasks, can significantly reduce the number of contexts and therefore the number of skills to learn. In the experiment shown in this work, this mechanism is able to identify 8 relevant contexts out of 512, thus also increasing the advantages of transfer learning. 

The experiments also shew how the framework of intrinsic motivations and the idea of ``artificial curiosity'' can be useful not only for task management, but also for identifying relevant contexts. Indeed, the surprise caused by an unexpected failure is a suitable trigger to bring the attention of the system to those environmental features that need to be taken into account to build new task-related policies.

Although the model allowed these achievements, we are aware that its current operationalisation has limitations that might be addressed in future work.
This could be facilitated by the fact that C-GRAIL can be considered as a blueprint architecture whose specific implementation and components can be modified and improved with computationally more effective components and additional mechanisms. 

A relevant limitation of the system are the perceptual components of the architecture.
In particular, we now arrange the system perception in terms of low-level features used for motor control and high-level features relevant for the acquisition of goals and the detection of contexts.
Future work might enhance the system by endowing it with the capacity of extracting relevant high-level features autonomously. To this purpose, for example, one might still divide the different use of proprioception for guiding control and images to guide the decisions on goals and contexts. However, for example, one might use auto-encoder neural networks to autonomously identify high-level meaningful features of objects as ``latent causes'' generating the images, encoded in the network ``bottleneck'' layers \cite{KingmaWelling2013}.
This would allow the autonomous identification of high-level features.

A problem related to the previous point is the fact that now when the agent fails to obtain the expected success for a goal due to the changed context the selection of the new relevant features (obstacles) is facilitated by the fact that the obstacles are discrete, their encoding is given to the agent as a vector of high-level features, and that the robot selects as relevant the features those that are close to the hand when a hit happens.
Future work might thus aim to allow the agent to autonomously guess which are the features that are actually relevant, for example by comparing the high-level features (e.g., extracted with an auto-encoder as discussed above) corresponding to the different contexts in order to focus on the different ones, and then by making different guesses and checks on which ones do actually matter. 

Last, testing the enhanced features of C-GRAIL discussed above would require the use of scenarios and tasks going beyond those considered here. For example, we required the learning of a simple reach-and-touch action to be performed on spherical objects; moreover, the environment contained only distinct and regular obstacles with same size and similar locations with respect to the target.  A more complex challenge might for example require the pursuit of more complex goals, such as picking up and placing objects in desired locations in a scenario containing several obstacles having variable size, shape, and spatial arrangement.

Notwithstanding these possible improvements, C-GRAIL represents an advancement with respect to previous systems for open-ended learning as it integrates for the first time two critical features needed for fully autonomous learning, namely the capacity to actively take into consideration contextual features only when they are relevant for the specific goal pursued and the capacity to adapt to the new relevant contexts through transfer learning.


%

\appendix[C-GRAIL: implementation details]
\label{sec:appendix}

This appendix provides the details about the implementation of C-GRAIL system which are not present in sec. \ref{sec:CGRAIL_implementation} . The sensory events that constitute the visual input to the system are encoded as an 80x60 binary (black and white) image determined by subtracting pixel by pixel two consecutive image frames provided by the camera of the right eye of the robot. When the binary image has at least one activated pixel, we consider that an event occurred (there has been a change in the environment) and its binary map is normalised (norm equal to 1), stored in the goal-representation map (GR-M), and associated (if new) to the first free unit in the goal selector. To prevent robot actuators from generating events with their movements we detect the arms and the scoops with a blue colour (not used for other elements of the setup) and when such colour is detected in pixels of either one of the two images used for change detection, we exclude the detection of change in correspondence to such pixels.


The contextual features that the smart context-detector (SCD) receives as input are encoded in a binary vector stating the presence (1) or absence (0) of each of the possible 6 obstacles: this input is compered through a Context Matching ($CM$) function to the lists of goal-specific useful contextual features $ucf^g$ to determine the current perceived context for each goal. When a \textit{failure} occurs, a new contextual feature might be added to $ucf^g$ if the current competence of the system on goal $g$ is over a certain $competence\_threschold$  set to 0.4.

The goal-specific contexts identified by the SCD (encoded in a binary vector with lenght equal to 7) are provided as input to the goal-selector that determines through a $softmax$ the current goal $g$ of the system. The probability $p(g|\phi^g)$ of $g$ to be selected given the goal-specific context $\phi^g$ is 
\begin{equation}
\label{eq:softmax}
p(g|\phi^g) =\frac{exp\left(\frac{Q(\phi^g,g)}{\tau}\right)}{\sum_{i=0}^{n}exp\left(\frac{Q(\phi^n,n)}{\tau}\right)} 
\end{equation}
where $Q(\phi^g,g)$ is the value of goal $g$ given context $\phi^g$, $n$ and $\phi^n$ are all the goals and their respective perceived contexts, and $\tau$ is the softmax temperature, set to 0.01, that regulate the stochasticity of the selection. At time $t$ the value of each goal $g$ is dependent on the current perceived contest for that goal $\phi^g$ and updated through an exponential moving average (EMA) of the intrinsic reinforcements provided by the compentence improvement $\Delta C^g_{\phi^g}$ obtained when practicing the goal in that context:
\begin{equation}
\label{eq:EMA}
    Q^{t+1}(g|\phi^g) = Q^t(g|\phi^g) + \gamma (\Delta C^g_{\phi^g} - Q^t(g|\phi^g))
\end{equation}
where $\gamma$ is a smoothing factor set to 0.3.
The competence-based intrinsic motivation (CB-IM) signal $\Delta C$ is determined by the activity of a competence predictor that evaluates the current (average) capability of the system to achieve goal $g$ given context $\phi^g$. In particular, the CB-IM signal is the difference between two averages of competence predictions ($CP$), where each average is calculated over an period $PT$ of 20 attempts (related to the same goal in the same context), so that the two averages cover a period of 40 attempts going backwards from the current selection into the past. Before covering the entire period $PTx2$, $\Delta C$ is calculated dividing by two the actual collection of predictions. In detail, at time $t$, the CB-IMs are calculated as follow:
\begin{equation}
\label{eq:IMs}
 \Delta C^g_{\phi^g} =  \frac{\sum_{i=t-(2PT-1)}^{t-PT}|CP{i}|} {PT}  - \frac{\sum_{i=t-(PT-1)}^{t}|CP_{i}|}{PT} 
\end{equation}

The expert-selector selects, on the basis of the selected goal and the current context for that goal, the expert that controls the robot during the trial. A softmax selection rule (eq. \ref{eq:softmax}) is used to determine the selected expert (with temperature set to 0.05), based on the values updated through an EMA (eq. \ref{eq:EMA}, with smoothing factor set to 0.3) of the matching signal generated for achieving the selected goal (1 for success, 0 otherwise).

The input to the selected expert is constituted by the angles of the four actuated joints of the (selected) arm of the robot, encoded through Gaussian radial basis functions (RBF) \cite{Pouget2000} with centres on the equally distributed vertexes of a 4 dimensional grid having 5 elements per dimension (the 5 units cover the range of one arm joint):
\begin{equation}
\label{eq:RBF}
y_i = e^{-\sum_d\left(\frac{\left(c_{d}-c_{id}\right)^2}{2\sigma_d^2}\right)}
\end{equation}
where $y_i$ is the activation of feature unit $i$, $c_d$ is the input value of dimension $d$, $c_{id}$ is the preferred value of unit $i$ with respect to dimension $d$ and $\sigma_d^2$ is the width of the Gaussian along dimension $d$ (where widths are parametrised so that when an input is equidistant, along a dimension, to two contiguous units, the activation of both the units is 0.5).

Each expert is a neural-network implementation of the actor-critic model \cite{Sutton1998} adapted to work with continuous states and action spaces \cite{Doya2000, Schembri2007}. The output of the critic ($V$) is a linear combination of the weighted sum of the input units:
\begin{equation}
\label{eq:critic}
V = \sum_i^{N}y_{i}u_{i} + b_V
\end{equation}
where $u_{i}$ is the weight projecting from input unit {i} and $b_{V}$ is the bias. The 4 output of the actor are determined by a logistic transfer function:
\begin{equation}
o_j=\Phi\left(b_j+\sum_{i}^{N}u_{ji}y_{i}\right)~~~~~\Phi(x)=\frac{1}{1+e^{-x}} 
\end{equation}
where $b_j$ is the bias of output unit $j$, $N$ is the number of input units, $y_i$ is the activation of input unit $i$ and $u_{ji}$ is the weight of the connection linking unit $i$ to unit $j$. 

To determine the motor commands $o^n_j$, we add noise to the activation of the relative output $o_j$. Since the desired position of the joints are modified progressively, using white noise would generate extremely little movements and the arm of the robot would explore a small region of the joints space. For this reason the noise ($n$) added to the output of the actor is generated with a normal Gaussian distribution with average 0 and standard deviation ($sd$) 2.0, and passed through an EMA with a smoothing factor set to 0.08. Moreover, to manage the exploration/exploitation problem \cite{Sutton1998} we implemented an algorithm that let the system autonomously regulate the noise $n$: the $sd$ of its dependent on a ``noise-decrease parameters'' ($d$) determined by an EMA (with smoothing factor set to 0.0005) of the success of the expert in achieving the goal for which it has been selected (1 for success, 0 otherwise), so that the higher is the competence of the expert, the lower the noise. The $sd$ of the selected expert $e$ at trial $T$ ($S_{eT}$) is updated as follow:
\begin{equation}
S_{eT} = sd(1-d)
\end{equation}
The actual motor commands are then generated as follows:
\begin{equation}
 o^n_j=o_j+n 
\end{equation} 
where the resulting commands are limited in [0; 1] and then remapped to the velocity range of the respective joints of the robot determining the applied velocity ($\dot{\alpha}$, $\dot{\beta}$, $\dot{\gamma}$, $\dot{\delta}$).

Actor-critic experts are trained through a TD reinforcement learning algorithm \cite{Sutton1998}. The TD-error ($\delta$) is computed as:
\begin{equation}
\delta = (r_t+\gamma_{k}V_t)-V_{t-1}  
\end{equation}
where $r_t$ is the reinforcement at time step $t$, $V_t$ is the evaluation of the critic at time step $t$, and $\gamma$ is a discount factor set to 0.99. The reinforcement is 1 when the robot achieves the selected goal, 0 otherwise. To speed up the exploration process of the robot, we introduced negative rewards (although they are not needed to learn the proper policies) for hitting obstacles (-1) and when trials ends without collisions (-0.5).

The weight $u_{i}$ of the critic input unit $i$ of the selected expert is updated as usual:
\begin{equation}
\Delta u_{i} = \eta_c \delta y_i 
\end{equation}
where $\eta_c$ is the learning rate, set to 0.02. The weights of the actor are updated as follows:
\begin{equation}
\Delta u^{a}_{ji} = \eta_{a} \delta (o^n_{j}-o_{j})(o_{j}(1-o_{j}))y_{i}  
\end{equation}
where $\eta_a$ is the learning rate, set to 0.4, $o^n_{j}-o_{j}$ is the difference between the control signal executed by the system (determined by adding noise) and the one produced by the controller, and $o_{j}(1-o_{j})$ is the derivative of the logistic function.

The \textit{learning threshold} to activate transfer learning process (sec. \ref{sec:TransferLearning}) is set to 0.4, while the \textit{transfer threshold} for the teaching expert is set to 0.7 and the \textit{transfer probability} is set to 0.5.

\section*{Acknowledgment}
This research was supported by the European Union’s Horizon 2020 Research and Innovation Programme under Grant Agreement No 713010, Project “GOAL-Robots – Goal-based Open-ended Autonomous Learning Robots”. 

\ifCLASSOPTIONcaptionsoff
  \newpage
\fi



\bibliographystyle{IEEEtran}
\bibliography{SantucciEtAl_MultiSkills_arxive}
\end{document}